# Co-Indexing Labelled DRSs to Represent and Reason with Ambiguities

Uwe Reyle

Institute for Computational Linguistics,
University of Stuttgart,
Azenbergstr. 12, 70174 Stuttgart

June 16, 1995

**Abstract**

The paper addresses the problem of representing ambiguities in a way that allows for monotonic disambiguation and for direct deductive computation. The paper focuses on an extension of the formalism of underspecified DRSs to ambiguities introduced by plural NPs. It deals with the collective/distributive distinction, and also with generic and cumulative readings. In addition it provides a systematic account for an underspecified treatment of plural pronoun resolution.

## 1  Introduction

Whenever humans process natural language sentences or texts, they build up mental representations that leave some aspects of their meanings underspecified. In particular so for all kinds of ambiguities, especially scope ambiguities of quantifiers and ambiguities that arise from the distributive/collective distinction of plural NPs. The mental representations we build up when we hear, or read ambiguous sentences cannot characterise the described situations more precisely than the sentences themselves. Only if additional information is available can such underspecified representations be refined towards partially (or even completely) disambiguated ones. But in almost all of the cases there is not enough information available to identify exactly one reading. (It is not even clear that the speaker of the sentence had exactly one reading in mind.) But nevertheless we may accept such sentences as true and will, therefore, use the underspecified representations as premises for our arguments. It is thus not enough to say what the underspecified representations look like and how they may be disambiguated. We also must be able to define a suitable consequence relation and to formulate inference rules for them.

The problem of reasoning with ambiguities is addressed in Poesio 1991, van Deemter 1991, Reyle 1993, and Reyle 1995. van Deemter 1991 considers lexical ambiguities and investigates structural properties of a number of consequence relations based on an abstract notion of coherency. He correctly rejects the idea of analysing ambiguous expressions as the disjunction of their disambiguations. Poesio 1991, Reyle 1993, and Reyle 1995 focus on quantifier scope ambiguities. Poesio 1991's inference schemata yield a very weak logic only; and Reyle 1993's deductive component is too strong. A systematic discussion of how to derive the consequence relation that holds for reasoning with ambiguities on the basis of empirically valid arguments is given in Reyle 1995. The consequence relation and the inference rules in Reyle 1995 reflect the fact that any occurrence of an ambiguous expression may/must be interpreted as dependent on some previous occurrence. This can be seen as coherency requirement, and thus is a point of contact with van Deemter's work. This dependency of interpretation will also play a crucial role in the present paper.

Note that inferences do not only play a role in arguments. There are inferences that come into play already during the interpretation process.

(1)  These five boys are terrible. At Hannah's and Lena's birthday they ate five cakes. They didn't do that at Kevin's birthday again.

Arguably the second sentence of (1) has some 20 readings. Nevertheless the (repetitive use of) **again** in the third sentence allows us to directly derive that the relevant birthday of Kevin's was after the mentioned birthdays of Hannah and Lena without explicitly considering all cases that correspond to different disambiguations. Another case of a lexically triggered inference is given in (2).

(2)  Fünf Softwarefirmen kauften dreizehn Computer. Anschließend liehen sie sie aus.
     Five software companies bought thirteen computers. Then they lent them out.

The German verb **ausleihen** is ambiguous. It can either mean **borrow** or **lend**. Assuming **ausleihen** to mean **borrow** leads to an inconsistency in the interpretation of (2), if the resultive state of the buying event, namely 'having x', is identified with the preconditions of **borrow**, i.e. 'not having x'. Detecting this inconsistency is again an inferential process that is independent of any particular disambiguation of (2). We will discuss this in detail in Section 6.

Let us further note that the verb **ausleihen** may not only be disambiguated without knowing in exactly which situations (2) is meant to be true, but also that the anaphoric links between the pronouns and their antecedent NPs are already established at this stage of interpretation. This poses a further constraint on the underspecified representations. They must be able to establish an anaphoric link between a plural pronoun and its antecedent NP already at a stage where syntactic and semantic plurality of the pronoun may still diverge. This is better explained by (3).

(3)  The lawyers hired a secretary they liked.

(3) has a reading according to which each of the lawyers hired a secretary

he liked. In this case the pronoun **they** is interpreted as semantically singular. Its syntactic plurality is licensed by the plurality of its grammatical antecedent. Semantically it is interpreted as individual variable bound by the distribution over the set of lawyers. We now claim that any link that is established between a pronoun and its grammatical antecedent should obey the monotonicity requirement, according to which the transition from an underspecified representation $r_1$ to a less underspecified (or even fully specified) representation $r_2$ is achieved only by *adding* information. I.e. the underspecified representations we are going to develop must provide some means to specify the impact of possible disambiguations of the antecedent phrase (in our example: the fact that the subject NP is interpreted distributively, or not) to the set of readings of the phrase containing the pronoun.

In the next section we give three reasons for labelling DRSs, that are related to the construction and representation of ambiguities. We then briefly show how scope ambiguities of quantifiers are dealt with in the theory of UDRSs. A new definition of the semantics for UDRSs is given in Section 3. This definition allows for a direct formulation of the consequence relation that respects aspects of coherence, i.e. may treat different occurrences of ambiguous phrases to mean the same thing (in each possible disambiguation of the text containing them). It also allows for the representation of dependent readings. This will be shown when the theory is extended to deal with collective/distributive ambiguities of plural NPs as well as with generic and cumulative readings they may license. Section 4 provides the basics of this extension. And sections 5 and 6 deal with the problem to link a plural pronoun to its antecedent phrase such that the property of monotonic disambiguation is preserved. Section 5 considers intra-sentential anaphoric relationships and inter-sentential links between a plural pronoun and a group of entities that is built by abstraction. Section 6 concentrates on dependent readings.

## 2 A Short Introduction to UDRSs

Before presenting the formalism of UDRSs I will state three further requirements that a language of underspecified meanings should meet. As the language of UDRSs directly implements these requirements it presents an approach to underspecification that is not only natural but also has advantages over approaches that do not fulfill them.

The first requirement states that we must be able to represent any partial order of scoping relations. Consider the sentences in (4). Because the scope of generalised quantifiers is clause bounded, [Everybody]$_1$ has wide scope with respect to [many a problem about the environment]$_2$ and [every politician]$_3$, while the relative scope of the latter two quantifiers is not determined in (4.a). In (4.b) only [most politicians]$_2$ must have narrow scope with respect to [Everybody]$_1$. In contrast to in (4.a) the interpretation of the third NP, i.e. the indefinite [a book on economics]$_3$, is not restricted to the clause in which it occurs.

(4) a. [Everybody]$_1$ believed that [many a problem about the environment]$_2$ preoccupied [every politician]$_3$.

   b. [Everybody]$_1$ believed that [most politicians]$_2$ had read [a book on economics]$_3$.

This shows that we need a representation language that is able to directly represent and manipulate *partial orders* of quantifier scope. The strategy to achieve an underspecified representation of quantifier scopings by partially instantiating the final order of scoping relations[1] is not suited to deal with sentences like (4.a) and (4.b). Such an approach would fix the relative scope of $Q_1$ and $Q_2$ in (4.b) by partially instantiating the final sequence of quantifier scope to $\langle Q_1, Q_2 \rangle$ and leave $Q_3$ 'in store' until there is enough information available to add it to the list. But note that inserting $Q_3$ amounts to imposing a linear order between $Q_1$, $Q_2$ and $Q_3$. There is thus no possibility to implement a weaker requirement saying that $Q_3$ has wide scope over $Q_2$. To be able to represent and monotonically disambiguate partial orders of scoping relations we must, therefore, give up the idea of dealing with scope ambiguities by such a kind of storage mechanism. We must directly talk about the partial relations, i.e. about pairs of quantifiers and not about a (set of) quantifier(s) and a (partially intstantiated) sequence. Saying that $Q_3$ has scope over $Q_2$ in our example, but enters no scoping relation with respect to $Q_1$, would then amount to extending the set $\{\langle Q_1, Q_2 \rangle\}$ to $\{\langle Q_1, Q_2 \rangle, \langle Q_3, Q_2 \rangle\}$. This requirement is fulfilled by the language of UDRSs and their construction procedure.

The second requirement concerns the representation of indefinite NPs. Within DRT indefinite NPs introduce discourse referents that may be bound by sentence constituents unrelated to the introducing phrase. This feature allows for a simple treatment of donkey sentences, like (5).

(5) Every student who owns a book on semantics reads it.

The semantic contribution of **a book on semantics** is a partial DRS of the form $\boxed{\begin{array}{c} x \\ \text{book-on-semantics}(x) \end{array}}$ . The quantificational force of the introduced discourse referent **x** is then determined by its position within the structure that represents the meaning of the whole sentence. If the indefinite is interpreted specifically then it ends up in the top DRS as in (6).

---

[1]This is essentially the idea in Alshawi and Crouch 1992 and HPSG (Pollard and Sag 1994).

(6) 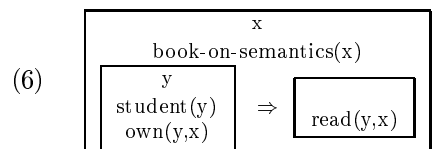

The universal interpretation of **a book on semantics** is shown in (7), where the discourse referent **x** is bound by the universally quantified NP **every student**.

(7) 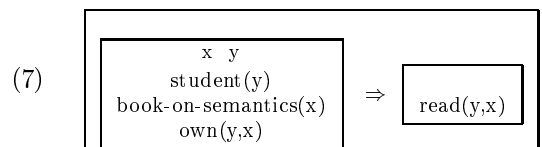

As the translations of (6) and (7) into predicate logic use two different types of quantifiers ($\exists x(... \wedge ...)$ in the case of (6) and $\forall x(... \rightarrow ...)$ in the case of (6)) there is no direct way of representing the meaning set of (5) in a single formula of some underspecified predicate logical language. In the Core Language Engine,[2] for example, so-called unresolved quantifiers are used to represent (5) by a quasi-logical form (QLF). QLFs do, however, not admit model-theoretic interpretation. In order to be interpretable the QLF for (5) must be resolved, i.e. translated into the two fully specified representations of standard predicate logic.

The third requirement states that the binding relations between NP meanings (i.e. between quantifiers, or discourse referents declared in some universe of a DRS) and their occurrences in subformulas should be preserved under any disambiguation. Algorithms that create all possible scoping relations between quantifiers (and operators) occurring in a single unscoped representation typically do not meet this requirement. To compensate this they use some meta-level constraint to rule out 'ill-formed' output. Consider (8)

(8)  Every professor who works with an industrial partner has at least two beautiful secretaries.

and the possible output (9)

(9)  $\forall x(professor(x) \wedge work\text{-}with(x,y) \rightarrow$
    $\exists^{\geq 2} z(b.secretary(z) \wedge \exists y(ind.partner(y) \wedge has(x,z))))$

of an algorithm that assigns the NP **an industrial partner** narrow scope with respect to **at least two beautiful secretaries** which in turn has narrow scope with respect to **every professor**. Then the so-called 'free-variable-constraint' (Pereira 1990, Hobbs and Shieber 1987) rules out (9) as a possible interpretation of (8) because the occurrence of $y$ in *work-with(x,y)* is free in (9). The need for such a meta-level constraint results from the fact that neither the unscoped representations themselves nor the disambiguation algorithm are subject to a corresponding demand on well-formedness. As a consequence the free-variable-constraint is not directly applicable to *partial* disambiguations. If we are to decide whether a partial disambiguation step is permitted, we must check whether the free-variable-constraint holds for each of the total disambiguations compatible with this step.

---

[2]See Alshawi 1992

Let us try to do better: Suppose we had some means to express in the object language that the formula *work-with(x,y)* (that we assume to be part of the unscoped representation for (8)) is a 'subformula of the second conjunct' of both, the translation $\lambda Q \exists y(ind.partner(y) \wedge Q(y))$ of **an industrial partner** and the translation $\lambda Q \forall x(professor(x) \rightarrow Q(x))$ of **every professor** – meaning that after $\lambda$-conversion for $Q$ ($y$ and $x$) the second conjunct in $\lambda Q \exists y(ind.partner(y) \wedge Q(y))$ and $\lambda Q \forall x(professor(x) \rightarrow Q(x))$ will turn into a formula that contains *work-with(x,y)* as a subformula. As these conjuncts correspond to the 'nuclear scope' of the quantification over $x$ and $y$ let us refer to them with the terms *scope(x)* and *scope(y)*, respectively. We then express the subordination relation by $work\text{-}with(x,y) \leq scope(x)$ and $work\text{-}with(x,y) \leq scope(y)$. Similarly we get $has(x,z) \leq scope(x)$, and $has(x,z) \leq scope(z)$, where $\lambda Q \exists^{\geq 2} z(b.secretary(z) \wedge Q(z))$ is the translation of **at least two beautiful secretaries**. Furthermore, we know from the syntactic analysis that $work\text{-}with(x,y)$ modifies the restrictor of **every professor**, which we express by $work\text{-}with(x,y) \leq res(x)$, where $res(x)$ denotes the antecedent of the implication in $\lambda Q \forall x(professor(x) \rightarrow Q(x))$.

Note that adding $\lambda Q \exists y(ind.partner(y) \wedge Q(y)) \leq res(x)$ would already correspond to a disambiguation step, because it forces the indefinite to have narrow scope with respect to the universal quantification. Similarly, the set of readings with **every professor** having wide scope over **at least two beautiful secretaries** is selected by demanding $\lambda Q \exists^{\geq 2} z(b.secretary(z) \wedge Q(z)) \leq scope(x)$. But note that we cannot add $\lambda Q \exists y(ind.partner(y) \wedge Q(y)) \leq scope(x)$, because this would imply that $work\text{-}with(x,y)$ is a subformula of $scope(x)$, which is not possible since we already have $work\text{-}with(x,y) \leq res(x)$.[3] Besides the axioms for partial orders, $\leq$ must, therefore, fulfil a further constraint guaranteeing that a subformula $A$ of $B$ can only be a subformula of $C$ in case $C$ is a subformula of $B$, or $B$ is a subformula of $C$ (in some disambiguation).[4] This constraint will be part of the definition of UDRSs below. The disambiguation procedure for UDRSs can, thus, be formulated such that it automatically guarantees wellformed output for any disambiguation step (without the need to go all the way to the set of total disambiguations this step allows).[5]

---

[3] We use the fact that for generalized quantifiers we always have $scope(x) \not\leq res(x)$ and $res(x) \not\leq scope(x)$.

[4] We do not consider branching quantification here.

[5] It also guarantees a proper treatment of the related example (10).

We now introduce the language of UDRSs. It follows from the discussion above that the relation, $\leq$, of being a subformula will play a pivotal role. With respect to DRSs $\leq$ matches with the subordination relation between DRSs (and DRS conditions). Let us consider the DRSs (13) and (14) representing the two readings of (12).

(12) Everybody didn't pay attention.

(13) 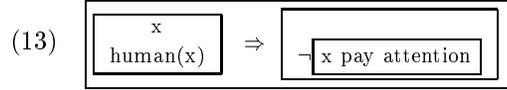

(14) 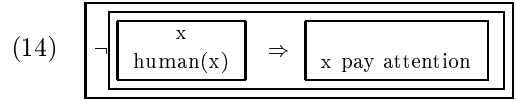

The following representations make the subordination relation – which is read from bottom to top – more explicit.

(15) 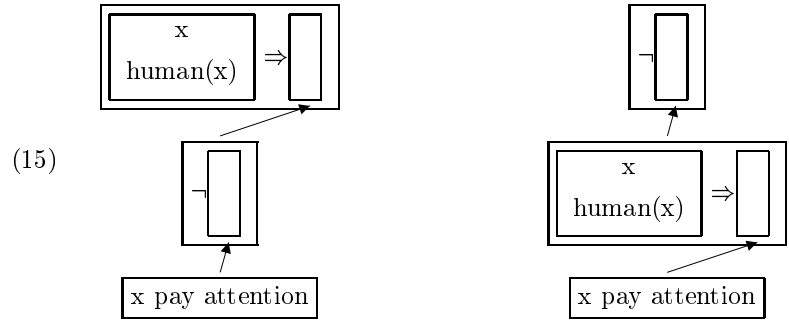

The structure that is common to both, (13) and (14), is represented by (16),

(16) 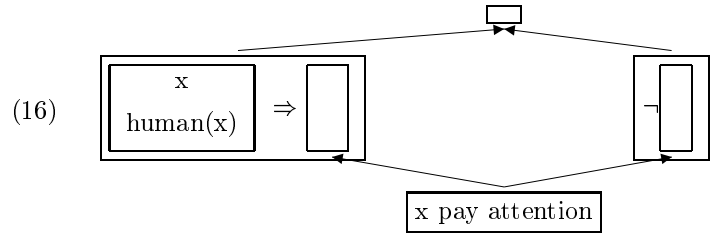

---

(10)   A manager of every company has a beautiful secretary.
And for examples involving pronouns, like
(11)   Every man saw a friend of his.
where the universally quantified noun phrase must have scope over the indefinite if the pronoun is assumed to be bound by the universal quantifier, the free-variable-constraint is implemented by the general principle saying that a bound pronoun must be bound within the scope of its binder. This principle can be expressed by a simple constraint of the form $l_\pi \leq l_\alpha$, where $l_\pi$ represents the meaning of the pronoun and $l_\alpha$ the scope of its antecedent. (For more details see Section 5.)

which is a graphical representation of the UDRS that represents (12) with scope relationships left unresolved. We call the nodes of such graphs UDRS-*components*. It is convenient to give each such component (in fact each sub-DRS occurring in a UDRS) a name, called its *label*. We furthermore define for every UDRS two functions, *scope* and *res*, which map labels of UDRS-components to the labels of their scope and restrictor, respectively. A UDRS consists of UDRS-components together with a partial order ORD of its labels. An example is given in (17).

(17) $l_\top : \langle \langle l_1 : [l_{11} : [x, human(x)] \Rightarrow l_{12} : [\;]], l_2 : [\neg l_{11} : [\;]], l_0 : [x\ pay\ att.] \rangle, ORD \rangle$

If $ORD$ in (17) is given as $\{l_2 \leq scope(l_1), l_0 \leq scope(l_2)\}$ then (17) is equivalent to (13), and in case $ORD$ is $\{l_1 \leq scope(l_2), l_0 \leq scope(l_1)\}$ we get a description of (14). If $ORD$ is $\{l_0 \leq scope(l_1), l_0 \leq scope(l_2)\}$ then (17) represents (16), because it only contains the information common to both, (13) and (14).

In any case ORD lists only the subordination relations that are neither implicitly contained in the partial order nor determined by complex UDRS-conditions. This means that (17) implicitly contains the information that, e.g., $res(l_2) \leq l_\top$, and also that $res(l_2) \leq l_2$, $res(l_1) \leq l_\top$, $scope(l_1) \leq l_\top$, and that neither $scope(l_1) \leq res l_1$ nor $res(l_1) \leq scope l_1$. We define

(18)  $l < l'$ iff $l \leq l'$ and not $l' \leq l$
      $l \sim l'$ iff $l \leq l'$ and $l' \leq l$

It is clear that disambiguation of UDRSs is a monotonic process. If we add $l_2 \leq l_{12}$ to (16) we get a representation equivalent to (13). There is thus no need to restructure (parts of) a semantic representation if more information about scope restriction has become available. This process of enrichment is characteristic for the construction of UDRSs: Information from different sources (syntactic[6] and semantic knowledge as well as knowledge about the world) may be incorporated in the structure by elaborating it in the sense just described.

The construction algorithm for UDRSs will associate meaning components of verbs with the lowest node of a UDRS-clause, sentence boundaries with its highest node and NP-meanings with the other nodes of the clause. For relative clauses the upper bound label $l'$ is identified with the label $l$ of its head noun (i.e. the restrictor of the NP containing the relative) by $l' \sim l$ (see clause (i.c.β) of the following definition). In the case of conditionals the upper bound label of subordinate clauses is set equal to the label of the antecedent/consequent of the implicative condition. The ordering of the set of labels of a UDRS builds an upper-semilattice with one-element

---

[6] An HPSG grammar for a fragment of German that realises these principles is presented in Frank and Reyle and Frank and Reyle 1995. Frank and Reyle focusses on scope ambiguities triggered by scrambling and/or movement.

$l_\top$. We assume that databases are constructed out of sequences $S_1, ..., S_n$ of sentences. Having a unique one-element $l_\top^i$ associated with each UDRS representing a sentence $S_i$ is to prevent any quantifier of $S_i$ to have scope over (parts of) any other sentence. The UDRS for (8) is given in (19).

(19) 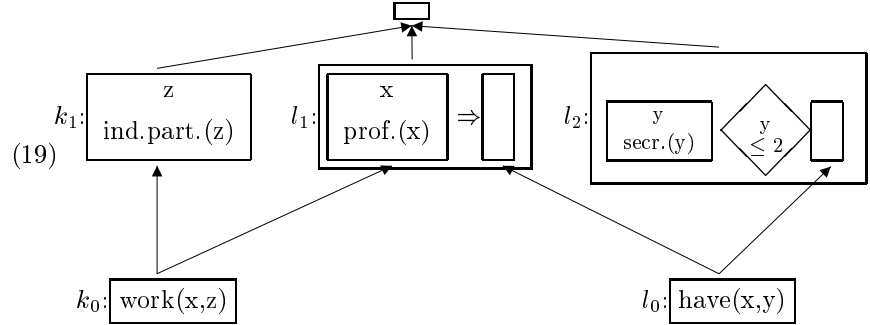

We see that adding $k_1 \leq scope(l_2)$ and $l_2 \leq scope(l_1)$ results in $k_0 \leq res(l_1)$ and $k_0 \leq scope(l_1)$, which – as discussed above – is a structure that doesn't correspond to a DRS in a natural way. (Recall that $\leq$ means nestedness of boxes.)

In clause (i.c.$\beta$) of the following definition we constrain the partial order of labels of a UDRS such that for implicative conditions and generalized quantifiers the set $\{res(l), scope(l)\}$ cannot have a lower bound. The definition of UDRSs furthermore ensures that

(i) the verb is in the scope of each of its arguments, (clause (ii.b)),

(ii) the scope of proper quantifiers is clause bounded, (clause (ii.c))

(iii) indefinite descriptions may take arbitrarily wide scope, (clause (ii.c) and (iii)).

**Definition 1:** Underspecified Discourse Representation Structures

(i) A quadruple $l:\langle\langle U_K, C'_K \cup C''_K\rangle, res(l), scope(l), ORD_l\rangle$ is a *UDRS-component* if the following conditions are satisfied:

  (a) $\langle U_K, C'_K \rangle$ is a DRS containing standard DRS-conditions only, and

  (b) $C''_K$ may contain at most one of the following conditions.

   ($\alpha$) a UDRS-clause $k:\gamma$ (defined under (ii) below),

   ($\beta$) $l_1:K_1 \Rightarrow l_2:K_2$,

   ($\gamma$) $l_1:K_1\langle Qx\rangle l_2:\langle\{\},\{\}\rangle$

   ($\delta$) $\neg l_1:\langle\{\},\{\}\rangle$

   where $K_1$ and $K_2$ are standard DRSs that may contain a UDRS-clause $k:\gamma$.

   The conditions in ($\beta$)-($\delta$) are called *distinguished* conditions of the UDRS-component, referred to by $l:\gamma$. UDRS-components with distinguished conditions are called *scope-bearing*.

- (c) *res* and *scope* are functions on the set of labels, and $ORD_l$ is a partial order of labels such that
  - ($\alpha$) if $k{:}\gamma \in C''_K$, then
    $k \sim l \in ORD_l$ and $ORD_k \subset ORD_l$;
  - ($\beta$) if $l_1{:}K_1 \Rightarrow l_2{:}K_2 \in C''_K$, or $l_1{:}K_1\langle Qx\rangle l_2{:}\langle\{\},\{\}\rangle$, then
    $res(l) = l_1$, $scope(l) = l_2$, and $l_1{<}l$, $l_2{<}l$, $l_1\not\leq l_2$, $l_2\not\leq l_1 \in ORD_l$;
    and there is no $l_0$ such that $l_0 \leq l_1 \in ORD_l$ and $l_0 \leq l_2 \in ORD_l$;
    if in addition if $k{:}\gamma \in C_{K_1}$, then $k \sim l \in ORD_l$ and $ORD_k \subset ORD_l$;
  - ($\gamma$) if $\neg l_1{:}K_1 \in C''_K$, then $res(l) = scope(l) = l_1$ and $l_1{<}l \in ORD_l$.
  - ($\delta$) Otherwise $res(l) = scope(l) = l$.
- (ii) A *UDRS-clause* is a pair of the form $l{:}\langle\langle\gamma_0,...,\gamma_n\rangle, ORD_l\rangle$, where $\gamma_i = l_i{:}\langle K_i, res(l_i), scope(l_i), ORD_{l_i}\rangle$, $0 \leq i \leq n$, are UDRS components, and
  - (a) $ORD_{l_i} \subset ORD_l$, for all $i$, $0 \leq i \leq n$
  - (b) $l_0 \leq scope(l_i) \in ORD_l$ for all $i$, $1 \leq i \leq n$
  - (c) $l_i \leq l \in ORD_l$ for all $i$, $1 \leq i \leq n$, for which $l_i{:}\gamma$ is a generalised quantifier.
  
  For each $i$, $1 \leq i \leq n$, $l_i$ is called a *node*. $l$ is called *upper bound* and $l_0$ *lower bound* of the UDRS-clause. Lower bounds neither have distinguished conditions nor is there an $l'$ such that $l'{<}l$.
- (iii) A UDRS-clause $l_\top{:}\langle\Gamma, ORD_{l_\top}\rangle$ is a *UDRS* if $ORD_{l_\top}$ is an upper semi-lattice with one-element $l_\top$.[7]
- (iv) A *UDRS-database* is a set of UDRSs $\{l^i_\top{:}\langle\Gamma, ORD_{l^i_\top}\rangle\}_i$.
  A *UDRS-goal* is a UDRS.

For the fragment without plurals UDRS-components that contain distinguished conditions do not contain anything else, i.e. they consist of labelled DRSs $K$ for which $U_K = C'_K = \{\}$ if $C''_K \neq \{\}$.

The dynamic aspects of indefinite NPs can be accounted for by a suitable extension of the definition of accessibility to UDRSs and UDRS databases. Note that although the definition of accessibility for DRSs is given in terms of subordination (i.e. $\leq$) it yields a weaker notion if it is applied to UDRSs. Take, for example, a sentence with two indefinites. With respect to accessibility on DRSs the discourse referent introduced by the first indefinite is accessible from the second, and vice versa. But if we apply the same definition to UDRSs neither the first one is accessible from the second nor the second from the first. This of course presupposes that the only NPs occurring in the sentence are the two indefinites. If in addition any quantified NP (or a negation) is present then mutual accessibility between the two indefinites requires that they both have wide scope over the quantifier (or the negation). We will call the relation we get when applying the definition of accessibility to UDRSs *weak accessibility*. And we will us the term accessibility for the appropriate extension of this notion.

---

[7]This means that for all nodes $k$ without a 'path' to $l_\top$ a condition $k \leq l_\top$ is added to $ORD_{l_\top}$.

## 3 Truth and Consequences of UDRSs

To define an underspecified representation to be *true* if one of its readings is true may be defensible – but is certainly not sufficient as a basis for a suitable definition of logical consequence. Consider (20),

(20)  If the students get £100 then they buy books. The students get £100. $\models$ The students buy books.

which shows that sentences in a discourse are often disambiguated in tandem, with the effect that the same disambiguating option is taken for them. Thus the meaning of the premise of (20) is given by (21b) not by (21a), where $a_1$ represents the first and $a_2$ the second reading of the second sentence of (20).

(21) a. $((a_1 \to b) \vee (a_2 \to b)) \wedge (a_1 \vee a_2)$

   b. $((a_1 \to b) \wedge a_1) \vee ((a_2 \to b) \wedge a_2)$

We will call sentence representations that have to be disambiguated similarly *correlated ambiguities*. And we will express such correlations by co-indexing. The types of ambiguities we will consider are lexical ambiguities, ambiguities triggered by plural noun phrases and quantifier scope ambiguities. Lexical ambiguities will be represented by ambiguous atomic DRS-conditions, quantifier scope ambiguities by the partial order of labels, and ambiguities triggered by plural noun phrases by a combination of both. If two occurrences of atomic DRS-conditions $\gamma$ and $\gamma'$ are coindexed then they express the same lexical meaning. This is straightforward. But what does it mean for UDRS-components and UDRS-clauses to be co-indexed, and under which circumstances is it possible to co-index them? Suppose $l{:}\gamma$ and $k{:}\delta$ are UDRS-clauses. And let us assume that coindexing is done with respect to their labels, i.e. $l^i{:}\gamma$ and $k^i{:}\delta$. As UDRS-clauses express scope ambiguities in terms of the partial orders of their labels the co-indexing should imply a certain isomorphism between the orders. This isomorphism must in addition map the labels of the grammatically corresponding nodes onto each other. Recall that we defined UDRS-clauses in terms of a list of UDRS-components (and not a set). Let us assume that the order of the elements in this list is canonical (for each verb). Then we can say that two UDRS-clauses may be co-indexed if the isomorphism between $ORD_l$ and $ORD_k$ also respects the canonical order of the arguments of the verb. As $ORD_l$ contains also the labels of UDRS-clauses contained in compo-nents of $l{:}\gamma$ it seems reasonable to require in addition that the embedded UDRS-clauses of $l^i{:}\gamma$ and $k^i{:}\delta$ must also be co-indexed.[8]

We will, therefore, assume that co-indexing is inherited from UDRS-clauses to URDS-components and from URDS-components to the conditions they contain.

An isomorphism $\omega$ between $ORD_l$ and $ORD_k$ that respects all canonical orders of sub-clauses of $l^i{:}\gamma$ and $k^i{:}\delta$ is called an *isomorphism between UDRS-clauses* $l{:}\gamma$ and $k{:}\delta$. Let $\overline{ORD_l}$ be the set of linear orders that extend $ORD_l$ such that $l{:}\langle\gamma, \overline{ORD_l}\rangle$ is a UDRS-clause; and let $c_l$ be a choice function on $\overline{ORD_l}$. Similarly for $ORD_k$. We then say that the choice function $c = c_l \cup c_k$ *respects the index* $i$ of the two UDRS-clauses $l^i{:}\gamma$ and $k^i{:}\delta$ (in symbols $c^i$), if the isomorphism between their orders is preserved. Let $I$ be a set of indices, and $c$ a choice function that respects all $i \in I$ (in symbols $c^I$). If $c^I(\Gamma)$ is denotes the disambiguation of $\Gamma$ triggered by $c^I$, then the consequence relation we assume underlies ambiguous reasoning is given in (23).

(23)    $\forall c^I(c^I(\Gamma) \models c^I(\gamma))$

We now give the definition of truth for UDRS-databases. We will use $M, f \models K$ to mean that the embedding function $f$ verifies the DRS $K$ in model $M$ according to the standard truth conditions of DRSs. FV($\underline{l}$) is the set of discourse referents $x$ that occur *free* in at least one labelled DRS $k{:}K$ with $k < l$, and such that $x \notin U_{K'}$ for all labelled DRSs $k'{:}K'$ that are accessible from $K$ and for which $k' < l$ holds. Finally, let $\mathcal{E}$ be the set of sets $e$ of embedding functions of a certain $K$ in a certain model $M$. We interpret labels by means of elements of $\mathcal{E}$. $\Lambda$ denotes the empty function.

**Definition 2:** Let $l$ be a label, $M$ a model, $f$ an embedding function defined on the set of discourse referents declared in DRSs $K'$ labelled by a label $k$ such that $l < k$, and $c$ a choice function on $\overline{ORD_l}$.

  (i) Suppose $l$ is the label of a lower bound, or occurs in a distinguished condition of some UDRS-clause. Let $K$ be the DRS labelled by $l$. Then $\|l\|^c_{f,M} = \{g \mid dom(g) = dom(f) \cup U_K \cup FV(\underline{l})$ and for all $\gamma \in C_K \ M, g \models^c_e \gamma\}$.

 (ii) Suppose $l$ is the label of a UDRS-component that is neither a lower nor an upper bound. Let $K$ be the DRS labelled by $l$. Then

---

[8]Consider the following variants of (20). If we assume that the pronoun **they** does not refer to the set of all students but to a subset of five of them the example shows that this notion of coindexing is too strong.

(22)    Five of my students will buy books if they get £100.

$$\left\{\begin{array}{l} \text{Five of my students} \\ \text{The students} \\ \text{They} \end{array}\right\} \text{get £100.} \models \text{They buy books.}$$

For the purposes of the present section it will, however, be sufficient. There are several directions in which the notion must be refined. One possible refinement will be discussed in a later section.

$\|l\|^c_{f,M} \in \mathcal{E}^{\mathcal{E}}$ such that $\|l\|^c_{f,M}(e) =$
$\{g \mid dom(f) \cup U_K \subseteq dom(g) \subseteq dom(f) \cup U_K \cup FV(\underline{l})$
and for all $\gamma \in C_K \ M, g \models^c_e \gamma\}$,
where $\models^c_e$ is defined as follows.
  (a) $M, g \models^c_e \gamma$ iff $g \in e$ and $M, g \models \gamma$
      if $\gamma$ is a standard DRS-condition
  (b) $M, g \models^c_e l_1{:}K_1 \Rightarrow l_2{:}K_2$ iff
      $\forall h$ ( if $h \in \|l_1\|^c_{g,M}$, then $\|l_2\|^c_{h,M} \cap e \neq \{\}$)
  (c) $M, g \models^c_e \neg l_1{:}K_1$ iff $\|l_1\|^c_{g,M} \cap e = \{\}$
  (d) $M, g \models^c_e l_1{:}\gamma$ iff $\|l_1\|^c_{g,M} \neq \{\}$ for upper bounds $l_1$

(iii) Suppose $l$ is an upper bound, i.e. $l$ labels a UDRS-clause
$l{:}\langle\langle l_0{:}\gamma_0, ..., l_n{:}\gamma_n\rangle, \mathrm{ORD}_l\rangle$. Let $\langle l_{j_0}, ..., l_{j_n}\rangle$ be the linear order of $\{l_0, ..., l_n\}$ that is induced by $c(\overline{ORD_l})$. Let $c_j$ be the restriction of $c$ to $\overline{ORD_{l_j}}$, for $j \in \{j_0, ..., j_n\}$. Then
$$\|l\|^c_{f,M} = \|l_{j_1}\|^{c_{j_1}}_{f,M}(...(\|l_{j_n}\|^{c_{j_n}}_{f,M}(\|l_0\|^{c_{j_0}}_{f,M}))..)$$
We refer to $\|l_{j_{r+1}}\|^{c_{j_1}}_{f,M}(...(\|l_{j_n}\|^{c_{j_n}}_{f,M}(\|l_0\|^{c_{j_0}}_{f,M}))..)$ by $e^c_{l_{j_r}}$, $1 \leq r \leq n$.

Applying this definition to the UDRS in (17) gives the following denotations for its labels. (We omit explicit reference to the model $M$, writing $\|.\|^{\Lambda}_{\Lambda}$ instead of $\|.\|^{\Lambda}_{\Lambda,M}$.)
$\|l_0\|^{\Lambda}_{\Lambda} = \{g \mid dom(g) = \{x\}$ and $g \models x \ pay \ attention\}$
$\|l_2\|^{\Lambda}_{\Lambda}(e) = \{g \mid dom(g) \subseteq \{x\}$ and $\|l_{21}\|^{\Lambda}_g \cap e = \{\}\}$
$\|l_{21}\|^{\Lambda}_{\Lambda}(e) = \{g \mid dom(g) \subseteq \{x\}$ and $g \in e\}$
$\|l_1\|^{\Lambda}_{\Lambda}(e) = \{g \mid dom(g) = \{\}$ and $\forall h($ if $h \in \|l_{11}\|_g$, then $\|l_{12}\|_h \cap e \neq \{\})\}$
$\|l_{11}\|^{\Lambda}_{\Lambda}(e) = \{g \mid dom(g) = \{x\}$ and $g \models human(x)\}$

**Definition 3:** Let $I$ be the set of indices $i$, $\{i_r\}_r$ be the set of occurrences of index $i$ in a *UDRS-database* $\mathcal{K} = \{\langle l^i_{\top}{:}\Gamma, ORD_{l^i_{\top}}\rangle\}_i$, $\{c_j\}_j$ a set of choice functions.
  (i) $\mathcal{K}$ is *true* in a model $\mathcal{M}$ with respect to $\{c_j\}_j$ if
      (a) there is an isomorphism $\omega_i$ between coindexed clauses that respects $i$ for all $i \in I$, and
      (b) $\bigcap_j \|l^j_{\top}\|^{c_j}_{\Lambda,M} \neq \{\}$.
  (ii) $\mathcal{K}'$ follows from $\mathcal{K}$, $\mathcal{K} \models \mathcal{K}'$, iff for all $\{c_j\}_j$ if $\mathcal{K}$ is true in $\mathcal{M}$ with respect to $\{c_j\}_j$, then $\mathcal{K} \cup \mathcal{K}'$ is true in $\mathcal{M}$ with respect to $\{c_j\}_j$.

As the disjunctions in (24.a) cannot be represented by co-indexed UDRSs, (24.a) is not a tautology, because not all of its readings are true. And as the sentences in (24.b) also lack a common index the inference in (24.b) does not hold, because **Everybody was awake** does not follow from both readings of the premise.

(24) a. $\models$ Everybody slept or everybody didn't sleep.
b. Everybody didn't sleep. $\models$ Everybody was awake.

This means that ambiguities in the data are interpreted by the disjunction of their readings and ambiguities in the goal by their conjunction – if there are no indices which correlate them. Therefore (25) holds only if the UDRSs that interpret the two occurrences of **Everybody didn't sleep** are co-indexed.

(25) Everybody didn't sleep. $\models$ Everybody didn't sleep.

Co-indexing different occurrences of ambiguities amounts to saying that they mean the same thing. This is especially the case if they are uttered in the same context of interpretation. But suppose that the first occurrence of, e.g., **The students get £ 100** in (20) is interpreted with respect to a context that is different from the context relevant to interpret its second occurrence (e.g. because there are 100 pages of text in between the two occurrences), then there is the possibility that they actually mean different things. In this case the interpreter does not establish a correlation between them.

## 4 Ambiguities triggered by plural NPs

Plural NPs bear a high potential for creating ambiguities. For one thing, many of them can be understood either as denoting a collection of individuals or quantifying over the members of that collection and thus give rise to the well-known collective/distributive ambiguity. But there are further possibilities for interpreting sentences with plural NPs. (26.a) and (26.b) are examples of so-called generic and shared responsibility readings, respectively.

(26) a. The children in this city thrive.
b. The guys in 5b have been cheating on the exam again.

These readings differ from the distributive reading in that they can be accepted as true even if not all members of the set denoted by the subject NP are in the extension of the predicate expressed by the VP (when it is interpreted as a predicate of individuals). To see that they differ from the collective reading for a similar reason consider (27).

(27) The girls gathered in the garden.

(27) has only a collective reading. This means that a predicate P is true of a group X, if every member of X contributes in some way or other to the fact that P is true of X. In (27) the contribution is the same for each girl and consists of having the property of going to the garden (eventually with the intention to meet the others). The generic and shared responsibility readings of (26) differ from the collective readings because they can be accepted as true even if not all members of the set denoted by the subject NP are in the extension of predicates that stand in such a relation to the VP. To specify the relevant relations is the task of lexical theory (and part of the specification of world knowledge). The task of UDRT is to provide an underspecified representation which subsumes all these readings.

Collective and distributive uses of a verb $\gamma$ are determined by the type of discourse referents $\gamma$ takes. The UDRS in (29), for example, represents the collective reading of (28). (Discourse referents of type group are represented by capital letters.)

(28)  The lawyers hired a secretary.

(29) 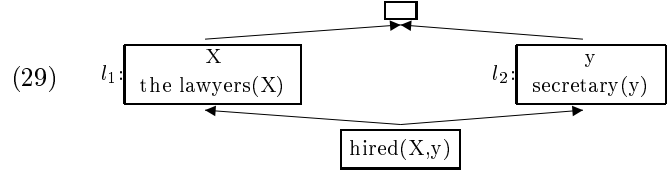

And its distributive reading is given in (30), where the quantification over the individual lawyers introduces a discourse referent, x, of type individual.

(30) 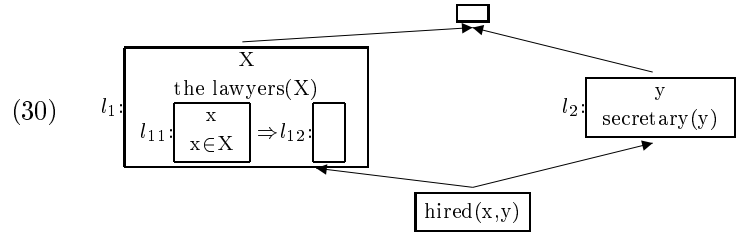

Let us note that while (29) is not ambiguous any more, the choice of the distributive reading (30) for (28) leaves leeway for a further ambiguity. This ambiguity is due to the fact that the node representing the subject NP has been turned into a scope-bearing node by applying distribution to **the lawyers(X)**. Thus the indefinite can be interpreted as being within the scope of the distribution, or not. In (29) the NP-node is not scope-bearing, and, therefore, the UDRS is equivalent to the DRS that results by taking the union/merge of all sub-DRSs of (29).

In order to come to a representation that is underspecified with respect to the choice of possible readings of (28), we first mark UDRS-components to which a distribution might still be applied as *potentially* scope-bearing, and second, leave it open whether the corresponding argument slot of the verb is instantiated by the plural discourse referent (in case the collective reading is chosen), or by the singular discourse referent that is bound by the distributive, generic or cumulative interpretation. To this end we simply use terms of the form $\alpha(\mathbf{X})$ to specify which NP occupies which argument slot of the verb. To indicate that the argument slot is filled by a plural individual we will add the condition $\alpha(\mathbf{X}) = \mathbf{X}$, and for the singular case we add $\alpha(\mathbf{X}) = \mathbf{x}$. To define the notion of potentially scope bearing, we modify the definition of UDRS-components such that *res* and *scope* are allowed to be partial functions. Thus clause (i.c.$\delta$) of Definition 1 must be restricted to labels $l$ for which *scope* and *res* are defined. It now reads: Otherwise $res(l) = scope(l) = l$, if $l \in dom(scope) \cap dom(res)$. And clause (ii.b) of Definition 1 is replaced by the following two clauses.

**(1.ii.b′)** $l_0 \leq scope(l_i) \in ORD_l$ for all $i$, $1 \leq i \leq n$, if $scope(l_i)$ is defined.
**(1.ii.b″)** $l_0 \leq l_i \in ORD_l$ for all $i$, $1 \leq i \leq n$, if $scope(l_i)$ is not defined.

**Definition 4:** Let **l** label a UDRS-component. Then
  (i) **l** is *scope bearing* if $scope(\mathbf{l}) \neq \mathbf{l}$.
  (ii) **l** is *not scope bearing* if $scope(\mathbf{l}) = res(\mathbf{l}) = \mathbf{l}$.
  (iii) if *res*, or *scope*, are not defined for **l**, then **l** is called *potentially scope bearing*.

Using this definition the underspecified representation (31) of (28) has exactly the shape of (29), but the argument DRSs $l_1$ is still marked as potentially scope bearing, i.e. *scope* and *res* are not yet defined for $l_1$.

(31)    $l_1$: [X | the lawyers(X)]    $l_2$: [y | secretary(y)]
        [hired($\alpha$(X),y)]

To disambiguate (31) we must not only add a condition of the form $\alpha(\mathbf{X}) = \mathbf{X}$, or $\alpha(\mathbf{X}) = \mathbf{x}$ (accompanied with 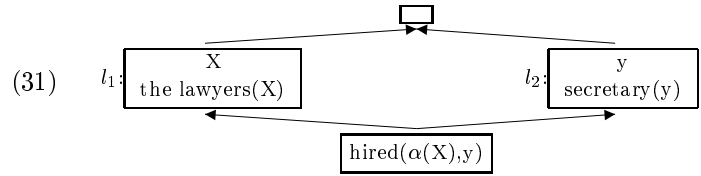), but also define *res* and *scope* for $l_1$. In case of the collective reading this will turn $l_1$ into a node that is not scope bearing. We thus define $res(l_1) := l_1$ and $scope(l_1) := l_1$. And in case the distributive reading is chosen we take $res(l_1) := l_{11}$ and $scope(l_1) := l_{12}$.

In a similar way the choice of a generic or shared responsibility reading can be dealt with. Both introduce a quantificational structure turning the node of the subject NP into a scope bearing one. The generic reading, e.g., for (26.a) may be represented by (32), in which GEN denotes the generic quantifier.

(32) 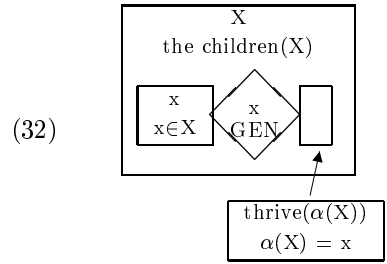

This method applies also to cumulative readings which are available when a verb is accompanied with two plural NPs, as in (33).

(33)  Three breweries supplied five inns.

Under the cumulative reading (33) can be accepted as true if for each of the three breweries there is at least one inn the brewery supplies, and each inn is supplied by at least one brewery. To represent this reading let us introduce a straightforward extension of (monadic) duplex conditions to "polyadic" ones. The restrictor of the polyadic duplex condition in (34) consists of the pair of DRSs associated with the nouns, and the diamond is not only equipped with the quantifications over the corresponding variables, but also marks the polyadic quantification to be cumulative by means of the superscript *cum*.

(34) 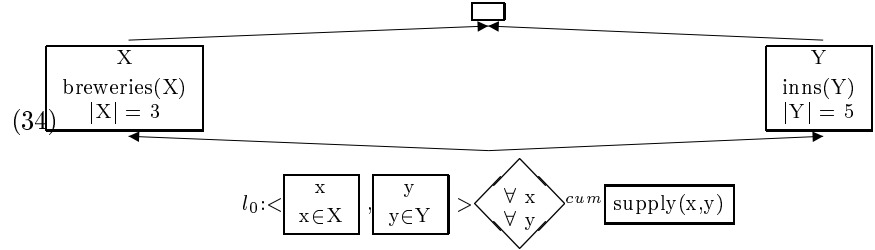

The verification conditions of the polyadic duplex condition in (34) are equivalent to those of the condition set in (35).

(35) 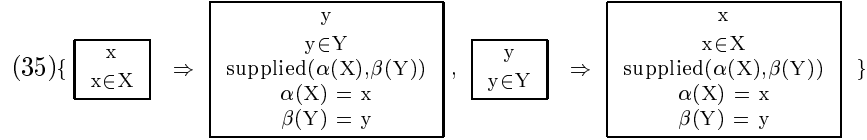

We now extend the verification conditions to intransitive and transitive verbs with underspecified argument types. We assume that the embedding function $f$ has $X$ (and $Y$) as well as $x$ (and $y$) in its domain.

(36)    - $f \models P(\alpha(X))$ iff $f \models P(X)$ or $f \models P(x)$
        - $f \models P(\alpha(X), \beta(Y))$ iff
            $f \models P(X,Y)$, or
            $f \models P(x,Y)$, or, $f \models P(X,y)$, or $f \models P(x,y)$, or

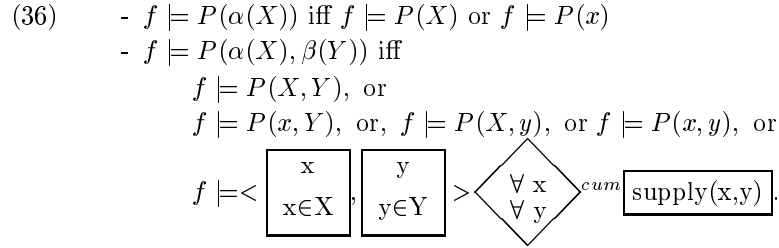

For UDRS-components introduced by plural NPs we assume that clause (ii) of Definition 2 applies only to labels $l$ for which *scope* is defined. We add the following clause to capture potentially scope bearing components.

(ii') Suppose $l$ is neither a lower nor an upper bound, and $scope(l)$ is not defined. Let $X$ be the distinguished discourse referent[9] of $U_K$. Then $\|l\|_f^c \in \mathcal{E}^{\mathcal{E}}$ such that $\|l\|_f^c(e) =$
$\{g \mid dom(f) \cup U_K \cup \{x\} \subseteq dom(g) \subseteq dom(f) \cup U_K \cup FV(l)$ and
$(g \models_e^c C_K$ or $(g \models^c C_K$ and $\{g \models_e^c \boxed{\begin{array}{c} x \\ x \in X \end{array}} \Rightarrow \boxed{\phantom{xx}} \})\}$.

For standard DRS-conditions $\models^c$ is the same as $\models$. Only for UDRS-clauses, i.e. upper bound labels, $\models^c$ is defined as $\models_e^c$ in Definition (2.ii.d). This accommodates relative clauses. Note that $g \models^c C_K$ does not entail $g \models_e^c C_K$: Because in the absence of the implicative condition the parameter $e$ constrains $g$ itself. And if the implicative condition is there, then $e$ constrains the scope of this condition.

Note furthermore that this clause does not affect the interpretation of indices associated with UDRS-clauses. To impose a constraint that co-indexed clauses must be interpreted in parallel we must guarantee that we choose embedding functions that enter the stage through the same disjunct in (36) above. The simplest way to achieve this is to interpret lower bound labels $l_0$ not as sets of embedding functions but as sets of pairs $\langle e, r \rangle$, where $e$ is a set of embedding functions that contains exactly those functions that verify the conditions of $l_0$ on the basis of one particular disjunct in (36) and $r$ says which disjunct this is. We will assume that $r$ takes one of the forms $c$, $d$, $\langle c, c \rangle$, $\langle d, c \rangle$, $\langle c, d \rangle$, $\langle d, d \rangle$, for the non-cumulative readings. And for cumulative readings we define $r = l_{02}$, where $l_{02}$ labels the scope of the polyadic duplex condition. The choice function that linearises ORD to compute the denotation of the upper bound label will then in addition to selecting a particular linearisation choose one element out of this set and record $r$ as particular choice associated with $l_0$.

What we have done for verbs may be extended to all lexical ambiguities. We, therefore, generalise clause (i) of Definition 2.

(2.i) Suppose $l$ is a lower bound, or occurs in a distinguished condition of some UDRS-clause. I.e. $l$ labels a standard DRS $K$. Suppose further that $\gamma$ is a lexically ambiguous expression in $K$, and that $\{K_1, ..., K_n\}$ is the set of meanings that results from $\gamma$'s ambiguity in $K$. Then $\|l\|_f^\Lambda = \langle \{g \mid dom(g) = dom(f) \cup U_{K_r}$ and $g \models C_{K_r} \}, r \rangle$, where $1 \leq r \leq n$.

We will assume that the output of the choice function is a pair $\langle lin, R \rangle$, where $lin$ is the particular linearisation chosen, and $R$ contains the information about the choices of particular lexical meanings for occurrences of ambiguous lexical expressions. This requires a straightforward modification of the definition of truth for UDRSs to guarantee that if several occurrences of an ambiguous lexical expression are co-indexed, then the isomorphism

---
[9]This is the discourse referent that represents the group described by the NP.

will also respect this kind of correlation by being the identity function on $R$.

## 5 The antecedents of 'They'

Plural pronouns are plural NPs and therefore share their ambiguity potential. In addition plural pronouns may be interpreted as 'individual variables', as shown by (37).

(37)  Few lawyers hired a secretary they liked.

The NP **few lawyers** in (37) binds the **they** in the embedded clause. Being a quantifying NP it does not introduce a plural discourse referent as possible antecedent for **they** and, therefore, has only one reading. This situation changes drastically if we replace the subject by a plural non-quantifying NP, as in (38).

(38)  The lawyers hired a secretary they liked.

If we ignore the cases where **a secretary** is interpreted specifically, then (38) has the five readings listed in (39).

(39) 
  a. $\exists X(lawyer(X) \wedge \exists y(secretary(y) \wedge hire(X,y) \wedge like(X,y)))$
  b. $\exists X(lawyer(X) \wedge \exists y(secretary(y) \wedge hire(X,y) \wedge \forall x(x \in X \to like(x,y))))$
  c. $\exists X(lawyer(X) \wedge \forall x(x \in X \to \exists y(secretary(y) \wedge hire(x,y) \wedge like(X,y))))$
  d. $\exists X(lawyer(X) \wedge \forall x(x \in X \to \exists y(secretary(y) \wedge hire(x,y) \wedge$
                                                 $\forall z(z \in X \to like(z,y)))))$
  e. $\exists X(lawyer(X) \wedge \forall x(x \in X \to \exists y(secretary(y) \wedge hire(x,y) \wedge like(x,y))))$

In (39.a) to (39.d) the pronoun is bound by the group variable X irrespective of the choice whether or not to quantify over the members of this group as in (39.c) and (39.d). Only in (39.e) this quantification binds the pronoun. It is interpreted as a singular bound variable, to which no further distribution is possible.

The previous section suggests (40) as underspecified representation of this set of five readings.

(40) 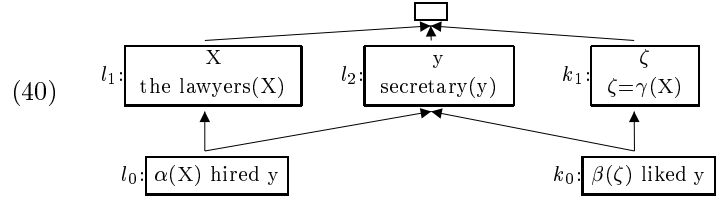

$\zeta$ is a neutral discourse referent, i.e. ranges over singular as well as plural entities. The anaphoric linkage between plural pronouns and their antecedent NPs is expressed by an equation of the form $\zeta = \gamma(X)$, where X is the discourse referent introduced by the antecedent NP and $\zeta$ the one introduced by the pronoun. The equation $\zeta = \gamma(X)$ leaves the choice to add $\gamma(X)=x$, or $\gamma(X)=X$.

(40) does, however, not automatically guarantee that all linearisations are proper, i.e. that the open variable problem doesn't occur. Take, for example, a variant of (38) in which **a secretary** is replaced by the quantifier **at least two secretaries**. Then it would be possible to linearise such that $l_1 < scope(l_2) < k_1$. We thus proceed as follows. The UDRS-components introduced by singular and plural pronouns are $l'$: [y; y=?] and $l'$: [ζ; ζ=α(?)], respectively. When resolving the pronoun we (i) instantiate '?', and (ii) add $l' < l$ to ORD, where $l$ is the label of ?'s value. This means that we have (41) instead of (40).

(41) 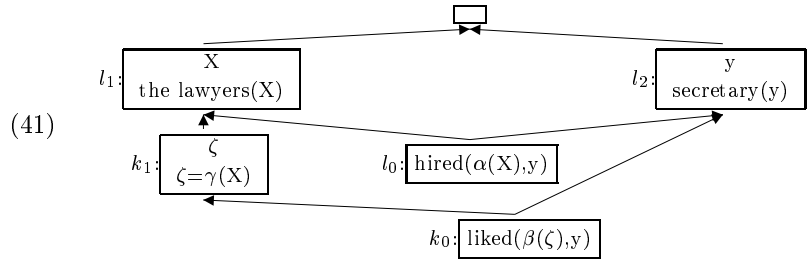

There are several ways to transform (41) via a couple of (partial) disambiguation steps into one of the fully specified readings given in (39). One possibility is to interpret **they** as referring to the group **X** of lawyers. This leaves the possibility of distribution open both with respect to **liked** and to **hired**. The different choices give us representations for (39.a) to (39.d). If we choose the collective reading of **hired**, then $α(\mathbf{X})=\mathbf{X}$ is added to $l_0$ in (41). We thus have deprived the subject UDRS of its scope bearing potential, because with $α(\mathbf{X})=\mathbf{X}$ also $scope(l_1) = l_1$ becomes part of the description of the UDRS. This blocks distribution with respect to **X**. Consequently, only $γ(X)=X$ can appear in $k_1$, and the only possibility for further disambiguating is to fix the interpretation of **liked**. Because $ζ$ is set equal to a group discourse referent it is possible to distribute over it. This yields (42) (= (39.b)).[10]

---

[10]Without further syntactic information (42) allows a disambiguation according to which the indefinite **a secretary** is within the scope of the distribution w.r.t. the verb **liked**. The problem of having more syntactic information available in order to restrict the disambiguation algorithm for UDRSs appropriately also shows up in the singular fragment and is discussed in Reyle 1993.

(42) 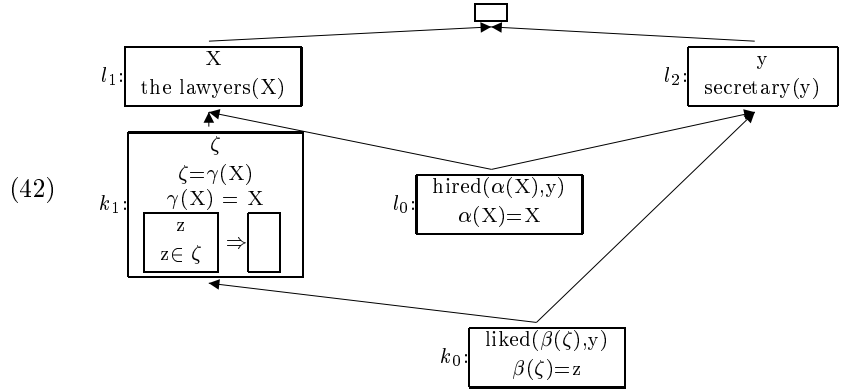

If the subject of **liked** is a collection, then no quantificational condition is introduced in $k_1$. We only add $\beta(\zeta){=}\zeta$ to $k_0$ and get the reading (39.a).

If we start by choosing the distributive reading of **hired**, (43) gets transformed into (41).

(43) 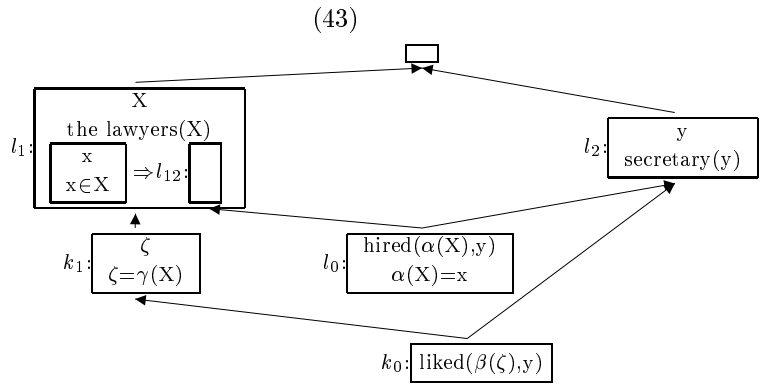

As the equation $\zeta{=}\gamma(X)$ still leaves open whether $\zeta$ is to be identified with **x** or **X** we now have the option to interpret **they** as bound by **x**. In this case we must add $k_1 < l_{12}$ to ORD and get (44), which represents the reading in (39.e).

(44)

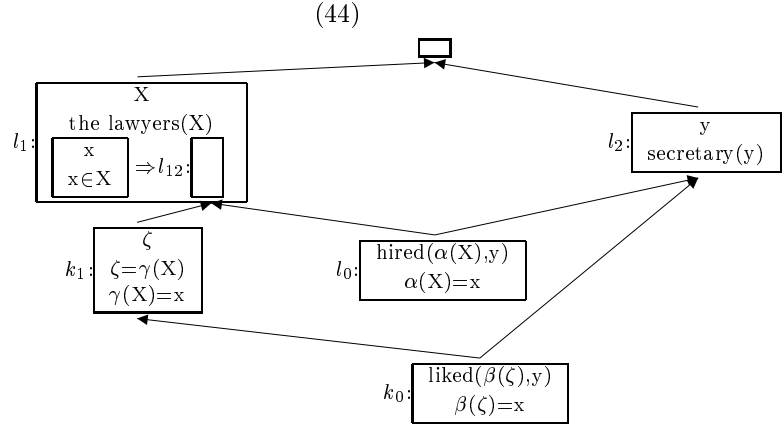

There are cases where no plural discourse referent has been introduced and the antecedent for **they** must be generated by Abstraction. Abstraction creates plural discourse referents by building the sum of discourse referents of DRSs that are created out of duplex conditions. This is done by, first, building the union K of the left hand and right hand side boxes of the duplex condition, and second, adding a condition of the form $\Sigma\eta$:K to the DRS in which the duplex condition occurs. To construct the antecedent of the **they** in (45), for example, we abstract over the discourse referent **x** in the DRS of the first sentence of (45) as shown in (46).

(45) Every teacher showed a picture to some child (in the class). They were bored.

(46) 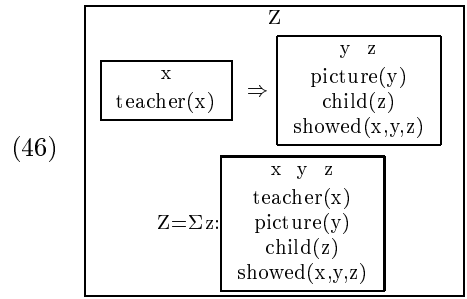

The discourse referent **Z** representing the group of children that have been shown a picture may now be picked up as antecedent for the pronoun. Note, however, that (46) represents only one reading of the first sentence of (45). There is also the reading where the indefinite **a picture** is interpreted specifically, as shown in (47).

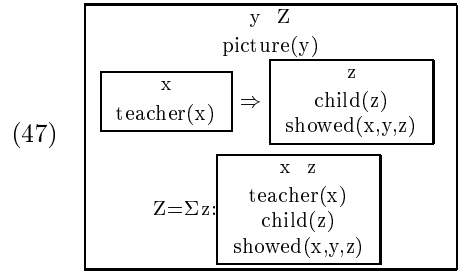

(47)

As **a picture** has wide scope over the universal quantifier its content doesn't show up in the DRS $K$ used for Abstraction. This ambiguity has to be preserved when Abstraction is applied to UDRSs. Consider the UDRS (48).

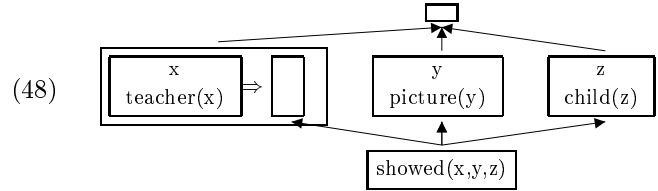

(48)

It not only represents the two readings (46) and (47) of (45), but also two other readings which we get out of (46) and (47) when we interpret **a child** specifically. Of course these readings are ruled out once we apply Abstraction to create a group of children. To establish the anaphoric link between the plural pronoun **they** in (45) and the NP **some child**, therefore, has disambiguating force. To implement this we must require that,

> whenever a plural pronoun is linked to an indefinite singular NP, then the link can only be established if there is some duplex condition occurring in the same clause as this NP. Furthermore the label of the indefinite NP must be set equal to the scope of one such duplex condition (whereas the label of the duplex condition should be accessible from the label of the pronoun).

We will call the duplex condition that supports the abstraction operation *licensing condition*. The link itself will be represented by a condition of the form $\zeta = \Sigma z{:}l$, where $z$ is the singular discourse referent introduced by the antecedent NP, and $l$ labels the licensing condition. This gives us the following as representation for (45).

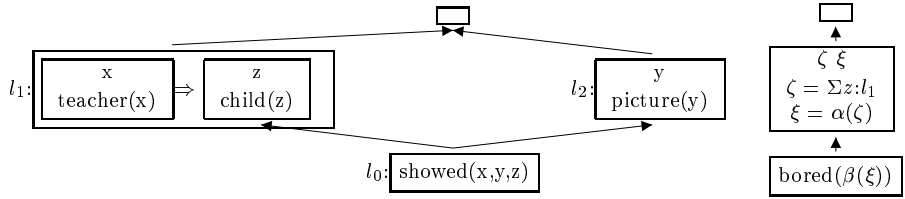

Standard DRT gives the following verification condition to equations of the form $Z = \Sigma z{:}K$ (see Kamp and Reyle 1993, p. 426).

(49) $\quad f \models_e^c Z = \Sigma z{:}K \quad \text{iff} \quad f(Z) = \oplus \{b \mid f \cup \{\langle z, b \rangle\} \models K\}$

Suppose the consequent of the licensing condition in (46) is labelled $l_{12}$. Then the set of embedding functions that verify the union of its restrictor and nuclear scope is equal to $\|l_{12}\|_g^c$. We thus define the verification condition for implicit Abstraction as follows.

**Definition 5:** Suppose $x$ is declared in a UDRS-component labelled $l_j$ belonging to some clause $l_{\top_i}{:}\langle\langle\gamma_0, ..., \gamma_n\rangle, \text{ORD}_i\rangle$, whose UDRS-components are labelled by $l_0, ..., l_n$. Suppose further that the licensing condition for $\xi = \Sigma z{:}l$ is $l_i{:}l_{i1} \Rightarrow l_{i2}$. Let $c_i$ be a linearisation of $\text{ORD}_i$. Then
$f \models_e^c \zeta = \Sigma z{:}l_i \quad \text{iff} \quad \forall e_{\underline{l_i}}^{c_i} \; f(\zeta) = \oplus\{b \mid f \cup \{\langle z, b \rangle\} \in \|l_{i2}\|_g^{c_i}(e_{\underline{l_i}}^{c_i})\}$[11]

This definition does, however, not cover so-called dependent uses of plural pronouns. (50) has a reading according to which **them** is interpreted as dependent on **they**, i.e. each child copied the picture that was shown to him.

(50)  Every teacher showed a picture to some child. They copied them.

To get this reading we cannot apply distribution with respect to both arguments of **copied** in the UDRS for the second sentence of (50). This would require far too many copyings of pictures by children. We must distribute over the subject and then interpret the object as dependent on the subject, in almost the same way we did in (44). The only difference is that in (44) the pronoun is *directly* bound by the quantification over the set of lawyers whereas in the case of (50) it is not the discourse referent introduced by distributing over the subject **they** which binds the object. Here the object **them** is bound *implicitly* in the following sense: Distribution over the subject not only amounts to considering all embedding functions $g$ that satisfy $\boxed{\begin{array}{c} w \\ w \in \zeta \end{array}}$. It really amounts to considering the functions $g$ satisfying $\boxed{\begin{array}{c} x \; y \; z \\ \text{teacher}(x) \\ \text{picture}(y) \\ \text{child}(z) \\ z = w \end{array}}$. In Kamp and Reyle 1993 this reinterpretation is achieved syntactically by accommodating the content of the DRS used for Abstraction within the restrictor of the duplex condition introduced by Distribution. We will not generalize this syntactic approach to the case under discussion. We will instead present a semantic solution.

---

[11] $e_{\underline{l_i}}^{c_i}$ is defined in Definition 2.iii.

## 6  Dependent Readings

The reading for (50) just discussed is a particular instance of *dependent* interpretations of verb meanings. Interpreting **copied** as dependent on **showed** in (50) means that the children-picture-pairs in the extension of **showed** are also in the extension of **copied**. Another case of dependent verb interpretation is (2) already mentioned in the Introduction, here repeated as (51).

(51)  Fünf Softwarefirmen kauften dreizehn Computer. Anschließend liehen sie sie aus.

We said that one cannot understand **ausleihen** in (51) to mean **borrow**. The reason being that the resultative state of the buying event cannot be consistently identified with the preconditions of **borrow**. Any justification of this relies on the fact that the interpretation of the second sentence of (51) is dependent on that of the first. To see this let us assume, for example, that a group of five software companies collectively bought thirteen computers. Then the second sentence of (51) can only mean that the group as such lent them out. It definitely cannot mean that each company borrowed the computers from the consortium (of which it is a member). Although this would not result in a contradiction, because each company does not possess any one of the computers on its own and, therefore, fulfills the preconditions of a borrowing event.[12] To exclude such independent interpretations we will mark the label $k_0$ of the second verb as dependent on the label of the first, i.e. $k_0^{dep(l_0)}$. This is to guarantee that in the case just described the collective reading of **kaufen** forces a collective reading of **ausleihen** (with respect to the subject NP). But note that the choice of the "same type" of reading is not sufficient. As the dependent reading on (50) showed we must interpret the coindexation as a constraint on the chosen sets of embedding functions[13] of the coindexed verbs (and not only as constraint on the choice itself). In the case of (50) we could have put the constraint into the verfication condition for the equation introduced by Abstraction. But consider (53), where no Abstraction is needed and the same dependent reading is possible. (We treat **didn't have any contract with** as transitive verb.)

---

[12]Intuitions might differ from ours for those who accept the sentences in (52).

(52)   Five software companies bought thirteen computers.
$$\left\{ \begin{array}{l} \text{Subsequently they borrowed them from the consortium.} \\ \text{Subsequently each of them borrowed one (from the consortium).} \end{array} \right\}$$

But even if the sentences in (52) are acceptable they are marginal, and the dependent interpretation of (51) is definitely preferred over the non-dependent one. Note that one may use **zurück/back** to force dependent interpretations. Like in **They immediately gave them back.**

[13]Recall that the denotation of a verb with underspecified arguments was a set of sets of embedding functions according to the modifiction (2.i) of Definition 2 on page 18.

(53)  Three breweries supplied$_{l_0}$ five inns. They [didn't have any contract with]$_{k_0^{dep(l_0)}}$ them.

As the second verb is marked dependent on the first[14] we will use this marking to constrain its extensions. As long as we don't know what the first sentence is supposed to mean, this constraint must apply for any possible disambiguation. So in case the first sentence is interpreted distributively with respect to both arguments the second should be interpreted in the same way. And in case the first has a cumulative interpretation the supply-relation must be included in the [didn't have any contract with]-relation. We have seen that whereas in the former case it is sufficient to say that the verb of the second sentence also has the distributive-distributive reading a correct interpretation is not ensured in the latter case simply by requiring a cumulative reading also for the second sentence. Here dependency marking must achieve more. It must make sure that any embedding function $f$ that verifies the nuclear scope of the polyadic duplex condition representing the cumulative reading of the first verb must also verify the scope of the condition for the second verb.

We proceed as follows. Suppose $l_0$ and $k_0$ are lower bound labels, such that $k_0$ is marked dependent on $l_0$, i.e. $k_0^{dep(l_0)}$. Let $FV(l)$ denote the set of discourse referents that occur (free or bound) in (some sub-DRS of the DRS labelled) $l$. We then restrict the set of embeddings that verify $k_0$ by those verifying $l_0$ as follows. Recall that $\|l_0\|$ is defined as sets of pairs on page 18.

**Definition 6:**

Suppose $\|l_0\| = \{\langle e_{l_0}, r_{l_0}\rangle\}_{r_{l_0}}$, $\|k_0\| = \{\langle e_{k_0}, r_{k_0}\rangle\}_{r_{k_0}}$, $\pi : FV(k_0) \mapsto FV(l_0)$. Then the restriction of $\|k_0\|$ to $\|l_0\|$ induced by $\pi$, short $\|k_0^{\pi:dep(l_0)}\|$, is $\{\langle e_{k_0}^\pi, r_{k_0}^\pi\rangle\}_{r_{k_0}}$, where

  (i)  $f \in e_{k_0}^\pi$ iff $f(x_i) = g(\pi(x_i))$ for some $g \in e_{l_0}^\pi$ and all $x_i \in FV(k_0)$
  (ii) $r_{k_0}^\pi = r_{l_0}^\pi$, in case $r_{k_0}^\pi \in \{c, d, \langle c, c\rangle, \langle d, c\rangle, \langle c, d\rangle, \langle d, d\rangle\}$
  (iii) $f \in \|r_{k_0}^\pi\|$ iff $f(x) = g(\pi(x))$ for some $g \in r_{l_0}^\pi$, in case $r_{k_0}^\pi$ labels the nuclear scope of $k_0$'s polyadic duplex condition.

To apply this to (51) let us assume that

  (i)  the resultative state of a buying event **e:kaufen(x,y)** is $\mathbf{s}_r^e$:**Have(x,y)**,
  (ii) the preconditions of a lending event **e** require $\mathbf{s}_p^e$:**Have(x,y)**, whereas the preconditions of a borrowing event require $\mathbf{s}_p^e$:¬**Have(x,y)**,

---

[14]We claim that situations like this occur very often in the interpretation process. Disambiguation requires a choice to be made among different concepts, lexicalized by some verb. In complex sentences, or texts, more than one verb occurs and each of them will be subject to this choice. More often than not, however, we cannot disambiguate. Nevertheless we are forced to either choose a specific concept type for all of these verbs simultaneously, or to allow a free choice to be made for each of them.

(iii) the meaning of **anschließend** triggers the identification of the resultative state of **kaufen** with the preconditions of **ausleihen**.

Then we get (54) as representation of the first sentence of (51).

(54) 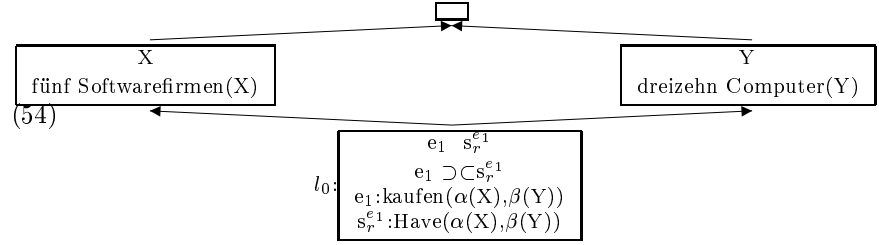

Without having disambiguated **ausleihen** the second sentence may be represented as in (55).

(55) 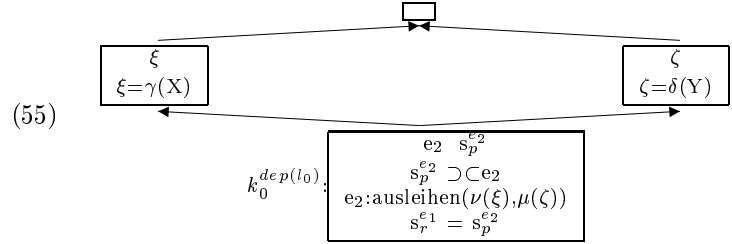

Now let us assume that **ausleihen** is meant in the sense of **borrow**, and that $\pi^{-1}(\alpha(X)) = \nu(\xi)$ and $\pi^{-1}(\beta(Y)) = \mu(\zeta)$. Then it follows from Definition 6 that

(56) $\quad f \in \|k_0^{dep(l_0)}\| \quad \text{gdw} \quad f \models$ $\begin{array}{c} e_1 \ s_r^{e_1} \ e_2 \ s_p^{e_2} \\ e_1 \supset \subset s_r^{e_1} \\ s_p^{e_2} \supset \subset e_2 \\ e_1:\text{kaufen}(\nu(\xi),\mu(\zeta)) \\ s_r^{e_1}:\text{Have}(\nu(\xi),\mu(\zeta)) \\ e_2:\text{ausleihen}_{borrow}(\nu(\xi),\mu(\zeta)) \\ s_p^{e_2}:\neg\text{Have}(\nu(\xi),\mu(\zeta)) \\ s_r^{e_1} = s_p^{e_2} \end{array}$

But this DRS cannot be verified at all, and therefore, **ausleihen** cannot mean **borrow** in (51). Note that we found the inconsistency in (56) (i) without having decided upon a particular interpretation of the first (and thus the second) sentence, and (ii) without considering the logical contribution of the arguments of **ausleihen**. Exactly the same reasoning will thus be applicable if we replace the object pronoun **sie** of (51) by **die meisten von ihnen** (meaning 'most of them').

## 7 Conclusion

We have presented an extension of the formalism of underspecified DRSs to deal with ambiguities triggered by plural NPs. We emphasized the fact that the representation language must be able to correlate different occurrences of ambiguous phrases, in the sense that any particular disambiguation chosen for one occurrence triggers the same disambiguation of all correlated occurrences. An even stronger requirement is given by dependent readings of verbs whose argument phrases contain plural pronouns. Here it is not only the same type of disambiguation that must apply to correlated verbs. The arguments of the dependent verb must be interpreted as bound by those of the one it is dependent on.

We introduced co-indexation and dependency marking to represent these types of correlated interpretations. And we have defined a semantics for them that does not presuppose any kind of accommodation to prepare the ground for dependent readings.

As a matter of fact indexation and dependency marking is also necessary in order to control reconstruction procedures for elliptical phrases, especially in cases of gapping (Fiengo and May 1994, Kamp ). As a matter of fact, the restrictions on reconstructing elliptical phrases and the restrictions on the disambiguation of dependent ambiguities are very similar. Results from the literature on ellipsis may, therefore, be used to refine the co-indexing mechanism used here. And the semantics given in the present paper may be generalised to apply to elliptical phrases. These are important questions for further research on underspecification.

## Acknowledgements

The first ideas of this paper where circulated as 'Monotonic Disambiguation and Plural Pronoun Resolution', DYANA-Report R.2.2.B, Oktober 1994. I am particulary grateful to Anette Frank, Esther König, Hans Kamp, Peter Krause, Larry Moss, and one anonymous reviewer, who gave comments on this report and on earlier versions of the present paper.